\documentclass[
]{ceurart}

\begin{document}

\copyrightyear{2021}
\copyrightclause{Copyright for this paper by its authors.
  Use permitted under Creative Commons License Attribution 4.0
  International (CC BY 4.0).}

\conference{Forum for Information Retrieval Evaluation, December 13-17, 2021, India}

\title{Pretrained Transformers for Offensive Language Identification in Tanglish}

\author[1]{Sean Benhur}[%
email=seanbenhur@gmail.com,
url=https://seanbenhur.github.io/,
]

\author[1]{Kanchana Sivanraju}[%
email= kanchana@psgcas.ac.in,
]

\address[1]{PSG College of Arts and Science, Civil Aerodrome Post, Coimbatore, India
  }

\begin{abstract}
This paper describes the system submitted to Dravidian-Codemix-HASOC2021: Hate Speech and Offensive Language Identification in Dravidian Languages (Tamil-English and Malayalam-English). This task aims to identify offensive content in code-mixed comments/posts in Dravidian Languages collected from social media. Our approach utilizes pooling the last layers of pretrained transformer multilingual BERT for this task which helped us achieve rank nine on the leaderboard with a weighted average score of 0.61 for the Tamil-English dataset in subtask B. After the task deadline, we sampled the dataset uniformly and used the MuRIL pretrained model, which helped us achieve a weighted average score of 0.67, the top score in the leaderboard. Furthermore, our approach to utilizing the pretrained models helps reuse our models for the same task with a different dataset. Our code and models are available in GitHub \footnote{\url{https://github.com/seanbenhur/tanglish-offensive-language-identification}}
\end{abstract}


\maketitle

\section{Introduction}
In the era of the internet, people from various age groups engage in social media, it has become a one-stop shop for all activities from learning to entertainment, but it is also filled with offensive and disturbing content, which is potentially harmful to everyone \cite{hegde2021images}. To prevent this, an automated system of identifying and flagging offensive content should be developed. Though there is a substantial amount of work done on major languages like English to identify offensive content \cite{chakravarthi2021dataset}, it is a challenging task to identify and flag offensive content in low resource languages, since many users tend to write their language in English script, which is called code-switching or code-mixing \cite{suryawanshi2021trollswithopinion,hande2021offensive,hande2021benchmarking}. Developing NLP systems on code-mixed text is challenging since the number of datasets is scarce\cite{HASOC-dravidiancodemix-2021,dravidiancodemix2021-overview,chakravarthi2021dravidiancodemix,chakravarthi2021dravidianmultimodality} and there are no clear patterns on these texts. The spelling and context might vary depending upon the user.

Dravidian languages are under-resourced in natural language processing \cite{chakravarthi2020leveraging}. Dravidian name was derived from Tamil, Dravidian means Tamil \cite{chinnappa-dhandapani-2021-tamil}, Dravidian languages are Tamili languages \cite{andrew-2021-judithjeyafreedaandrew}.  Tamil is a language spoken by Tamils in Tamil Nadu, India, Sri Lanka, and the Tamil diaspora worldwide, with official recognition in India, Sri Lanka, and Singapore \cite{sarveswaran2021thamizhimorph,b-a-2021-ssncse,b-a-2021-ssncse-nlp}. Current Tamil script was developed from the Tamili script, the Vatteluttu alphabet, and the Chola-Pallava script. There are 12 vowels, 18 consonants, and 1 ytam in this word (voiceless velar fricative) \cite{sajeetha9342640,sajeetha9185369,sajeetha9063341,thenmozhi2018ontology}. The Tamil script is also used to write minority languages including Saurashtra, Badaga, Irula, and Paniya. Tamil Eluttu "means" sound, letter, phoneme" in Tolkappiyam (about 5,320 BCE), and thus includes the sounds of the Tamil language, as well as how they are created (phonology) \cite{sakuntharaj2016novel,sakuntharaj2017use,sakuntharaj2018detecting,sakuntharaj2018refined}. All the Tamili (Dravidian) languages evolved from Tamil language \cite{r-c-n-2019-building,subalalitha2019information}. 

HASOC2021: Hate Speech and Offensive Content Identification is a competition that helps increase research in offensive language identification in code mixed languages such as  Tamil-English and Malayalam-English \cite{HASOC-dravidiancodemix-2021}. The dataset consists of comments/posts that were collected from Youtube and social media. Each comment/post is annotated with an offensive language label at the comment/post level. This dataset also has class imbalance problems depicting real-world scenarios.

In this paper, we present our system developed for HASOC 2021; the rest of the paper is organized as follows. Section 2 discusses the research work on offensive language identification and natural language processing in under-resourced languages. Following this, in section 3, we present the methodology of our work, from preprocessing the dataset, our model architecture, and training procedures. In section 4, we discuss our results. Finally, in section 6, we conclude with a summary and future work.

\section{Related Work}
Offensive Language identification has been widely made across many people in multiple languages. Shared tasks like HASOC-19 \cite{inproceedings} dealt with hate speech and offensive language identification in Indo-European languages. HASOC-Dravidian-CodeMix - FIRE
2020 \cite{chakravarthi2020overview}\cite{mandl2020overview} is the first shared task for identifying offensive content in Tamili languages. Previous work on Tamili languages on hope speech \cite{chakravarthi-2020-hopeedi,chakravarthi-muralidaran-2021-findings}, troll meme detection \cite{suryawanshi-chakravarthi-2021-findings}, multimodal sentiment analysis \cite{chakravarthi2021dravidianmultimodality} have paved the way to research in Tamili languages.

Researchers have used a wide variety of techniques for the identification of offensive language. There have been previous work \cite{lakshmanan2020theedhum}  in using classical machine learning models with efficient feature generation. Other researchers in \cite{vasantharajan-thayasivam-2021-hypers} \cite{sai-sharma-2021-towards} have used an ULMFit model \cite{ulmfit}  and pretrained XLM-Roberta model with translated and transliterated texts for this task.

\section{Methodology}

This section briefly describes our methodology for this task, including data preparation, model architecture, and training strategies. For this HASOC 2021 competition, we only use the datasets that were provided for the HASOC task. Table 1 shows the statistics of the train and dev distribution.

\subsection{Dataset}

The dataset given for subtask,  Offensive Language Identification in Tamil-English, consists of Youtube comments, present in code-mixed data containing text written in both native and roman scripts in English.

For training our model, we concatenate both training and dev sets; we remove the URLs, English stopwords, @username mentions, NAN values, emojis, and also punctuations; this preprocessing method is applied to all the train, dev, and test sets. After the task deadline, we sample the dataset uniformly to handle the class imbalance problem in this dataset, which helps us improve our score. Table 1 shows the statistics of the given dataset after preprocessing.

\begin{table*}
  \caption{Distribution of Tamil English Dataset}
  \label{tab:freq}
  \begin{tabular}{ccl}
    \toprule
    Distribution&Data\\
    \midrule
    Train & 4937\\
    Test & 1000\\
  \bottomrule
\end{tabular}
\end{table*}

\subsection{Model Architecture}
We use pretrained transformer models with custom pooled output for this task of identifying offensive content. We have used mBERT and MuRIL pretrained models from huggingface checkpoints. In this section, we describe our pooling operations on the pretrained models and the pretrained models.

\textbf{Attention Pooler}:
In this method, the attention operation described in the below equation is applied to the last hidden state of the pretrained transformer; empirically, this helps the model learn the contribution of individual tokens. Finally, the returned pooled output from the transformer is further passed to a linear layer for the prediction of the label.
\begin{equation}
   o = W^T_{h} softmax(qh^T_{CLS})h_{CLS}
\end{equation}
where $W^T_{h}$ and $q$ are learnable weights.
\begin{equation}
    y = softmax(W^T_{o} + bo)
\end{equation}

\textbf{Mean Pooler}:
In this method, the average of the last layer of the pretrained transformer is taken. This acts like a pooling layer in a convolutional neural net. An alternative to this method is to use max pooling, but max-pooling selects only the words with essential features rather than utilizing all the words. Since our dataset is code-mixed and the spelling of the tokens are not precise, we choose to go with mean pooling approach.

\textbf{mBERT}
Multilingual models of BERT \cite{devlin2019bert}. This model was pre-trained using the same pretraining strategy that was
employed to BERT, which is Masked Language Modeling (MLM) and Next Sentence Prediction (NSP).
It was pretrained on the Wikipedia dump of top
104 languages. To account for the data imbalance
due to the size of Wikipedia for a given language,
exponentially smoothed weighting of data was performed during data creation and word piece vocabulary creation. This results in high resource languages being under-sampled while low resourced
languages being over-sampled.

\textbf{MuRIL}
MuRIL \cite{khanuja2021muril}, pretrained model is trained on 16 different Indian Languages; the model was pretrained on Masked Language Modeling(MLM) and Translated Language Modelling(TLM). This model outperforms mBERT on all the tasks in XTREME \cite{hu2020xtreme}

\subsection{Training}
Though finetuning transformers gives better results and is dominant across leaderboards of various NLP competitions. Transformer models are still unstable due to catastrophic forgetting \cite{mosbach2021stability}. For this offensive language identification task, we carefully choose our hyperparameters for experimentation.  We finetune our custom models with binary-cross-entropy loss and AdamW optimizer, which decouples the weight decay from the optimization step. Linear scheduler for learning rate scheduling with 2e-5 as an initial step is used with this training strategy. The training hyperparameters are listed in Table 2.

\begin{table}[hbt!]
\caption{Hyperparameters used across experiments}
\begin{tabular}{lll}
\hline
\textbf{Hyperparameters} & \textbf{Values} &  \\ \hline
Learning Rate            & 2e-5            &  \\
Maximim Sequence Length  & 512             &  \\
Batch Size               & 8               &  \\            
Epochs                   & 5               &  \\
Weight Decay             & 0.01            &  \\
Dropout                  & 0.5             &  \\ 
AdamW epsilon            & 1e-06           & 
\end{tabular}
\end{table}

\section{Results and Discussion}

In this HASOC 2021 competition, the teams were ranked by the weighted F1-score of their classification system. This section discusses our experimental results; since we have used both training and dev sets for training, the train set in the dataset distribution refers to the concatenated given train and dev sets. The W-Precision, W-Recall, and W-F1-Score refer to the Weighted precision, weighted recall, and weighted F1-Score. Table 3 shows our results obtained before the task deadline using Attention Pooler and mBERT without sampling the dataset. After the task deadline, we uniformly sample our dataset and run or experiments on MuRIL and mBERT with  Attention Pooling and Mean Pooling. The results are provided in Table 4 and Table 5. The W-precision, W-Recall and W-F1 Score stands for Weighted Precision, Weighted Recall and Weighted F1-Score.

\begin{table}[hbt!]
\caption{Results before task deadline}
\begin{tabular}{@{}llll@{}}
\toprule
\textbf{Dataset Distribution} & \textbf{W-Precision} & \textbf{W-Recall} & \textbf{W-F1 Score} \\ \midrule
Train                         & 0.90                 & 0.91              & 0.92                \\
Test                          & 0.61                 & 0.60              & 0.61                \\ \bottomrule
\end{tabular}
\end{table}

\begin{table}[hbt!]
\centering
\caption{Results on Train dataset}
\begin{tabular}{@{}llll@{}}
\toprule
\textbf{Model}             & \textbf{W-Precision} & \textbf{W-Recall} & \textbf{W-F1 Score}  \\ 
\midrule
mBERT with AttentionPooler & 0.93                 & 0.90              & 0.93                 \\
mBERT with MeanPooler      & 0.90                 & 0.92              & 0.91                 \\
MuRIL with AttentionPooler & 0.88                 & 0.88              & 0.88                 \\
MuRIL with MeanPooler      & 0.93                & 0.93              & 0.93                 \\
\bottomrule
\end{tabular}
\end{table}

\begin{table}[hbt!]
\centering
\caption{Results on Test data}
\begin{tabular}{@{}llll@{}}
\toprule
\textbf{Model}             & \textbf{W-Precision} & \textbf{W-Recall} & \textbf{W-F1 Score}  \\ 
\midrule
mBERT with AttentionPooler & 0.65                 & 0.65              & 0.65                 \\
mBERT with MeanPooler      & 0.61                 & 0.61              & 0.61                 \\
MuRIL with AttentionPooler & 0.63                 & 0.63              & 0.63                 \\
MuRIL with MeanPooler      & 0.67                & 0.67              & 0.67                 \\
\bottomrule
\end{tabular}
\end{table}

From the above results, we conclude that the pretrained model MuRIL with MeanPooler performs best than others. Also, one can infer that the difference between training and test scores shows that the model is suffering from overfitting, and also sampling the dataset uniformly is a crucial step to increasing the score.

\section{Conclusion}

In this paper, we have presented our solution for the Offensive Language Identification system, which uses pretrained transformers mBERT and MuRIL. As a result, we achieve Rank 9 on the leaderboard and a 0.67 f1-score after the task deadline. For future research, we will consider improving the results by using any external dataset and other pretrained models and reducing the generalization error of the model.

\bibliography{sample-ceur}

\end{document}